\RequirePackage{tikz}

\documentclass[sn-apa,iicol]{sn-jnl}

\usepackage{amsmath}
\usepackage{amsthm}
\usepackage{amssymb}
\usepackage{booktabs}
\usepackage[switch]{lineno}
\usepackage{array}
\usepackage{multirow}
\usepackage{arydshln}
\usepackage{makecell}
\usepackage{dsfont}
\usepackage{pgfplots}
\pgfplotsset{compat=1.18}

\jyear{2022}%

\theoremstyle{thmstyleone}%
\theoremstyle{thmstyletwo}%

\theoremstyle{thmstylethree}%

\begin{document}

\title[Article Title]{Universal Object Detection with Large Vision Model} 

\author[1,2]{\fnm{Feng} \sur{Lin}}\email{lin1993@mail.ustc.edu.cn}

\author[1]{\fnm{Wenze} \sur{Hu}}\email{windsor.hwu@gmail.com}

\author[3]{\fnm{Yaowei} \sur{Wang}}\email{wangyw@pcl.ac.cn}

\author[3]{\fnm{Yonghong} \sur{Tian}}\email{yhtian@pku.edu.cn}

\author[2]{\fnm{Guangming} \sur{Lu}}\email{luguangm@hit.edu.cn}

\author[2]{\fnm{Fanglin} \sur{Chen}}\email{chenfanglin@hit.edu.cn}

\author[4]{\fnm{Yong} \sur{Xu}}\email{yxu@scut.edu.cn}

\author*[1]{\fnm{Xiaoyu} \sur{Wang}}\email{fanghuaxue@gmail.com}

\affil[1]{\orgname{Intellifusion Inc.}, \orgaddress{\city{Shenzhen}, \country{China}}
}
\affil[2]{\orgname{Harbin Institute of Technology, Shenzhen}, \orgaddress{\country{China}}
}
\affil[3]{\orgname{Peng Cheng Laboratory}, \orgaddress{\city{Shenzhen}, \country{China}}
}
\affil[4]{\orgname{South China University of Technology}, \orgaddress{\city{Guangzhou}, \country{China}}}


\abstract{
Over the past few years, there has been growing interest in developing a broad, universal, and general-purpose computer vision system. Such systems have the potential to address a wide range of vision tasks simultaneously, without being limited to specific problems or data domains. This universality is crucial for practical, real-world computer vision applications. In this study, our focus is on a specific challenge: the large-scale, multi-domain universal object detection problem, which contributes to the broader goal of achieving a universal vision system. This problem presents several intricate challenges, including cross-dataset category label duplication, label conflicts, and the necessity to handle hierarchical taxonomies. To address these challenges, we introduce our approach to label handling, hierarchy-aware loss design, and resource-efficient model training utilizing a pre-trained large vision model. Our method has demonstrated remarkable performance, securing a prestigious \emph{second}-place ranking in the object detection track of the Robust Vision Challenge 2022 (RVC 2022) on a million-scale cross-dataset object detection benchmark. We believe that our comprehensive study will serve as a valuable reference and offer an alternative approach for addressing similar challenges within the computer vision community. The source code for our work is openly available at \url{https://github.com/linfeng93/Large-UniDet}.}


\maketitle

\section{Introduction}\label{sec1}
A universal, general-purpose computer vision system has become a trend in the development of computer vision technology~\citep{he2022x,wang2019towards,gong2021mdalu,hasan2021generalizable,zhou2022simple}. This universal vision model is a multi-talented agent that can simultaneously solve a wide range of vision tasks with minimal human intervention. Researchers no longer need to train separate models for each individual vision task or fine-tune an existing model for a specific data domain. Instead, they can achieve all tasks with a single effort. The universality of this model is a promising direction towards human-like AI and has important implications for real-world computer vision applications~\citep{yuan2021florence}.

In this study, we aim to contribute to the development of universal vision technology. Specifically, we focus on the large-scale universal object detection problem across different domains. The goal is to have a single object detector that can perform the inference process once and generate unified detection results across all datasets, regardless of their differences.

The challenge of developing a large-scale multi-domain universal object detection system lies in two areas: (\emph{a}) curating a large-scale and diverse training dataset, and (\emph{b}) creating a robust visual representation method.
The training dataset must cover a wide range of data domains in order to achieve satisfactory results across domains. However, such a dataset is currently not available.
Furthermore, building a unified large-scale dataset with dense annotations for object detection~\citep{zhao2020object,zhou2022simple} and similar fine-grained computer vision tasks~\citep{lambert2020mseg,bevandic2022automatic,ranftl2020towards} is cost-prohibitive.
In terms of robust visual representations, it is challenging to ensure the common object detector is robust to a vast and diverse source data, often numbering in the millions, as objects of interest can vary greatly in different images.


The increasing availability of object detection datasets has opened the door to implementing universal object detectors by repurposing these resources. Our strategy involves consolidating these datasets by harmonizing their distinct label spaces, enabling us to tackle multi-domain object detection tasks characterized by varying label vocabularies. However, the integration of multiple diverse datasets often introduces annotation inconsistencies, including label duplication, conflicts, and incomplete hierarchical taxonomies. To overcome these challenges, we have developed a comprehensive loss formulation within the unified label space, ensuring the robustness of our approach.

In addressing the robustness challenge inherent in the multi-domain object detection problem, our strategy revolves around harnessing the power of large pre-trained vision models.
Recent studies have demonstrated the superiority of larger models in capturing higher-quality visual representations compared to smaller models~\citep{radosavovic2020designing,bello2021revisiting,kolesnikov2019revisiting}. These high-quality representations lead to better generalization both in-domain and out-of-domain~\citep{goyal2022vision}. Thus, we believe that the use of large well-trained vision models would significantly improve the performance of universal object detection for million-scale diverse datasets. Our experiments indeed show a noticeable improvement in performance by using larger vision models in the universal object detection task.

As we acquire the feature robustness by taking advantage of large pre-trained vision models, computational resources become a critical demand because of both computational and memory-wise costs. 
Without many computational resources yet, we introduce a resource-efficient training formulation for large vision models inspired by a recent work~\citep{vasconcelos2022proper}, which saves considerable computational resources, especially GPU memory, during the training procedures.

This paper presents our novel approach to the challenge of multi-domain universal object detection at a scale of millions of diverse images. We utilize the power of large pre-trained vision models and present an efficient training formulation that saves computational resources. Our method, Large-UniDet, has achieved remarkable performance and won the \emph{second} prize in the object detection track of the Robust Vision Challenge 2022 (RVC 2022)\footnote{\url{www.robustvision.net/leaderboard.php?benchmark=object}, IFFF\_RVC entry on Leaderboard}. The success of our approach is attributed to the efficient formulation design, careful label handling, and knowledge transfer from large-scale pre-training. 

\begin{figure*}[!t]
  \centering
  \includegraphics[width=6.3in]{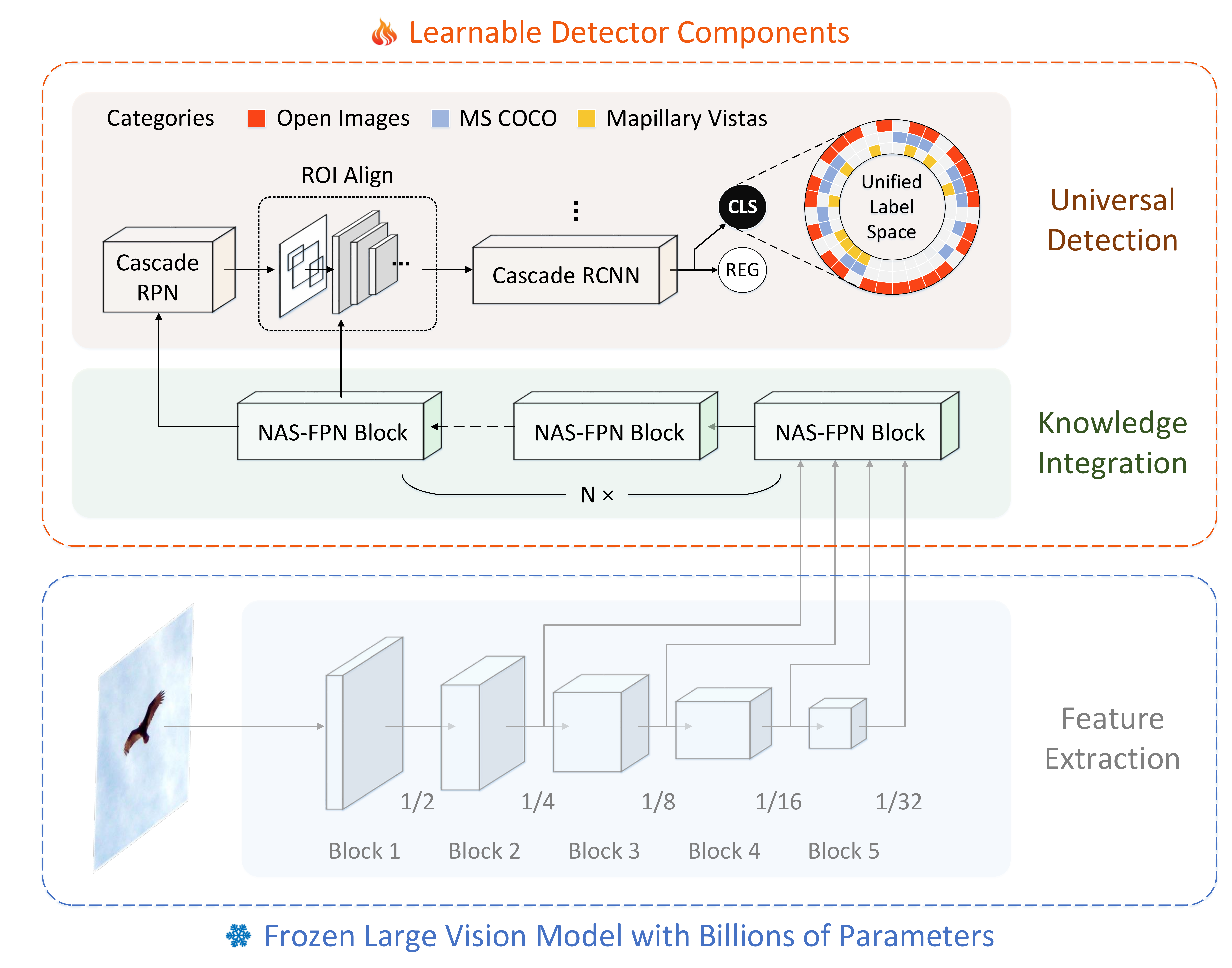}
  \caption{Overview. The design of Large-UniDet is based on a two-stage RCNN-style object detection network. The frozen backbone is a RegNet architecture initialized with the weights of SEER models. The NAS-FPN blocks can be stacked N times for a better accuracy-cost trade-off. The classification branch of Cascade R-CNN outputs 541 class scores including the \emph{background} as the cardinality of the unified label space is 540.}\label{fig:framework}
\end{figure*}

Our contributions are summarized as follows.
\begin{itemize}
\item Our approach explores the use of large vision models for the challenging task of large-scale multi-domain universal object detection.
\item We present a resource-efficient training formulation for large vision models in universal object detection, which saves computational resources during training procedures.
\item With the unified label space, we handle multi-domain object detection with different label vocabularies and overcome cross-dataset label duplication and semantic hierarchy problems.
\item The proposed method, Large-UniDet, achieved the $2nd$ prize in the object detection track of RVC 2022, demonstrating its impressive performance and robustness.
\end{itemize}

\section{Related Works}
\subsection{Universal Object Detector}
Recent years have seen a growing interest in universal object detection. Wang et al.~\citep{wang2019towards} propose a universal detector with a domain attention module that leverages shared knowledge across different data domains. The design consists of multiple dataset-specific detectors that share most network parameters while keeping the categories of each dataset separate.
Universal-RCNN~\citep{xu2020universal} proposes a different approach by incorporating graph transfer learning for modeling both intra-domain and inter-domain semantics of categories drawn from multiple datasets~\citep{lin2014microsoft,krishna2017visual,zhou2017scene}.
Unlike these methods, Zhao et al.~\citep{zhao2020object} build a unified label space by manually merging multiple label spaces of different datasets, and their framework is dedicated to managing partial annotations through the use of pseudo-labeling. UniDet~\citep{zhou2022simple}, in contrast, presents an automatic method to unify label spaces based on visual concepts generated by a partitioned object detector with three separate branches. Cai et al.~\citep{cai2022bigdetection} construct a unified label space by extracting category embeddings from each dataset using a language model. Recently, Meng et al.~\citep{meng2022detection} leveraged pre-trained language embeddings to generate adapted queries for each category embedding across different datasets, modeling object classification as a region-word alignment problem without a merged label space.

Similar to the methods mentioned above based on the unified label spaces, we propose a solution to improve the unified label space in universal object detection by modifying the manually-crafted taxonomy used in the RVC challenge. Our proposed method also addresses the challenges posed by label duplication and semantic hierarchy issues across multiple datasets.

\subsection{Pre-training for Vision Tasks}\label{ref-pretrain}
Pre-training is a widespread technique in computer vision~\citep{sun2017revisiting,joulin2016learning,kornblith2019better,caron2020unsupervised,cai2022bigdetection,azizi2021big} that enhances performance by using backbones models trained on large-scale datasets such as ImageNet~\citep{deng2009imagenet}, JFT-300M~\citep{sun2017revisiting}, OpenImages~\citep{kuznetsova2020open}, or web-collected data~\citep{goyal2022vision}. The backbone generates robust visual representations that can benefit various downstream vision tasks~\citep{goyal2022vision}. For object detection, the choice of the pre-trained backbone is crucial for determining performance~\citep{liu2020cbnet}. Typically, the strength of a pre-trained backbone comes from its (\emph{a}) powerful architecture, (\emph{b}) broad training data, and (\emph{c}) advanced pre-training methods.

\emph{Stronger architecture.}
To explore the influence of backbone architectures on object detection performance, Huang et al.~\citep{huang2017speed} examined the correlation between backbone capacities and performance. In contrast, Liu et al.~\citep{liu2020cbnet} enhanced the backbone's capabilities by amalgamating multiple identical backbones.
Furthermore, Liu et al.~\citep{liu2022swin} leveraged the power of vision transformers to create an extremely large object detector by using an expanded Swin transformer~\citep{liu2021swin} as the feature extractor. 

\emph{Training data.} 
In terms of the training data, Sun et al.~\citep{sun2017revisiting}  have demonstrated the impact of using a large-scale dataset JFT-300M on the robustness of representation learning. Bu et al.~\citep{bu2021gaia} have taken a different approach by combining various detection datasets~\citep{shao2019objects365,kuznetsova2020open,lin2014microsoft,dollar2011pedestrian,zhang2017citypersons} to attain better pre-trained weights for transfer learning in downstream tasks. Meanwhile, Cai et al.~\citep{cai2022bigdetection} have utilized existing detection datasets~\citep{gupta2019lvis,shao2019objects365,kuznetsova2020open} to create a large pre-training dataset through careful curation based on well-defined principles. 

\emph{Pre-training approach.} 
Recent advancements in backbone pre-training ~\citep{caron2020unsupervised,he2020momentum,he2022masked,lin2021auto,xu2022seed} have demonstrated the superiority of self-supervised methods over supervised approaches in computer vision tasks, such as object detection, semantic segmentation, and image classification. With the advantage of utilizing unlimited diverse image data from the web, self-supervised pre-training methods are capable of capturing more discriminative visual representations without relying on vast manual annotations~\citep{goyal2022vision}. Furthermore, the training of vision foundation models on large-scale image-text data~\citep{yuan2021florence,jia2021scaling,radford2021learning,shao2021intern} highlights the significant impact of representation learning on both in-domain and out-of-domain downstream tasks.

To enhance the performance of our universal object detector, we have chosen to use large \textbf{SE}lf-sup\textbf{ER}vised vision models known as SEER models~\citep{goyal2022vision} as the backbone. As discussed earlier, robust backbones can be attributed to high-capacity architectures, a diverse training data, and cutting-edge pre-training techniques. The largest SEER model that we will be using boasts a massive 10 billion network parameters. The SEER models are trained on a self-supervised clustering-based method~\citep{caron2020unsupervised} utilizing 1 billion less biased uncurated images collected from the web. This results in robust visual representations that perform well on both in-domain and out-of-domain benchmarks~\citep{goyal2022vision}. SEER has exhibited greater robustness, fairness, reduced harm, and minimized bias when compared to lots of supervised models. Our belief is that the SEER backbones will be capable of producing more discriminative features and provide better out-of-distribution generalization for the task of universal object detection across datasets with varying characteristics.

\subsection{Vision-Language Systems}
Prior to the advent of universal vision models, universal language models had already achieved remarkable success in the field of natural language processing (NLP)~\citep{howard2018universal}. The emergence of large language models (LLMs) has brought about a profound transformation in traditional paradigms for NLP tasks~\citep{radford2018improving, devlin2018bert, raffel2020exploring}. These models employ a unified approach, leverage billion-scale multi-source corpora data for training, and operate within a consistent tokenized framework for various NLP tasks.

Given the remarkable success of language models, the research community is now turning its attention to achieving vision universality, working on the development of language-assisted, multi-skilled general-purpose vision systems capable of addressing a wide range of vision tasks. Gupta et al.~\citep{gupta2022towards} introduce a General Purpose Vision (GPV) system based on a vision-language architecture. GPV exhibits the ability to learn and perform tasks involving image input, producing textual descriptions or bounding boxes. Building upon this concept, Kamath et al.~\citep{kamath2022webly} have devised an effective method to scale GPV by learning skills from supervised datasets, acquiring knowledge from web image search, and leveraging GPV's capacity to transfer visual knowledge across diverse skills. Furthermore, Unified-IO~\citep{lu2022unified} incorporates language model technologies that standardize input and output representations into a common vocabulary token format across all tasks. This approach allows a single model to excel in a wide range of vision tasks and vision-and-language tasks, consequently amplifying the versatility of these systems. These works demonstrate that leveraging extensive training data from multiple domains, diverse types, and various modalities offers a feasible approach to achieving a general-purpose AI agent.

\section{Method}

Our method can be effectively divided into two distinct components.
The first part meticulously outlines our object detection framework, highlighting specialized designs that enhance the efficient training of large vision models.
The second part delves into practical strategies employed for joint training across multiple diverse specific datasets.
These strategies are presented in comprehensive detail and can be readily adapted to analogous scenarios involving other datasets.

\subsection{Resource-efficient Detection with a Large Vision Model}\label{network}

In this section, we introduce our strong object detector that is built on large pre-trained backbone networks. The use of large vision models has been demonstrated to improve the performance of many computer vision tasks. However, the enormous computational and memory requirements for training these models limit their practical use~\citep{dai2021coatnet,shao2021intern,radford2021learning}. To address this challenge, we propose a computationally \& memory efficient training approach that freezes the parameters of the billions of pre-trained backbone neurons and fine-tunes the extracted visual representations on the subsequent detector components. 
This enables training of our largest model using a restricted GPU configuration of 16 NVIDIA 3090 GPUs (detailed in~\ref{training_strategies}), reducing the GPU memory consumption by approximately fourfold (NVIDIA A100/H100-80G GPUs may be required otherwise).
Our resource-efficient approach leverages the recent advancements in knowledge transfer~\citep{vasconcelos2022proper} and is specifically designed for large pre-trained vision models, providing a valuable resource for the computer vision community that is interested in object detection with limited computational resources.

Fig.~\ref{fig:framework} illustrates the overall framework. Each detector component is described in detail in the remaining content of this section.

\subsubsection{Frozen Backbone with Billions of Parameters}\label{network:backbone}



We provide a concise overview of the backbone employed in our proposed object detector, followed by an introduction to the advantages of our frozen design. As highlighted in Section~\ref{ref-pretrain}, our decision to employ the SEER models~\citep{goyal2022vision} as the backbone of our object detection network stems from their remarkable ability to enhance fairness and mitigate bias across a spectrum of diverse domains. This strategic selection ensures the consistent extraction of robust visual representations across three quite distinct datasets: MS COCO (COCO), Mapillary Vistas Dataset (MVD), and OpenImages Dataset (OID).



Traditionally, optimizing an object detector involves fine-tuning both the initialized backbone and subsequent detector components using detection datasets. Nonetheless, fine-tuning the backbone on smaller datasets can lead to the drift of the backbone parameters from their pre-trained initialization, potentially undermining detection performance~\citep{vasconcelos2022proper}. Furthermore, fine-tuning a resource-intensive backbone can significantly escalate computational complexity. To attain exceptional detection performance while managing computational demands, we have opted to maintain the frozen state of the backbone parameters throughout the training process. This strategic approach not only conserves resources but also contributes positively to the performance of long-tailed object categories through knowledge preservation~\citep{vasconcelos2022proper}. This preservation holds significance within the realm of the RVC multi-domain scenario.

Drawing from the observation that enhanced in-domain and out-of-domain generalization is positively correlated with the scale of the SEER model, as demonstrated by Goyal et al.~\citep{goyal2022vision}, we conduct a meticulous assessment of the trade-off between cost and performance. This evaluation guides us in selecting the optimal configuration for our experiments and, subsequently, our final submission to the RVC competition.
In the end, we opt to use both the lighter version (SEER-RegNet32gf) and the second largest version (SEER-RegNet256gf) for our extensive evaluations.

\subsubsection{Cascade Detection Heads}\label{network:detector}
In order to enhance the performance of our object detector, we implemented a two-stage RCNN-style detection framework with the frozen SEER backbone. Initial experiments using Faster R-CNN~\citep{ren2015faster} did not produce satisfactory performance, likely due to its limited number of learnable parameters making it difficult to handle the large-scale detection tasks across diverse datasets. Taking inspiration from recent advances in the knowledge transferring field~\citep{vasconcelos2022proper}, we adopted a high-capacity Cascade R-CNN~\citep{cai2018cascade} as our detection heads, which greatly improved detection performance as discussed in Section~\ref{exp:det_components}.

\subsubsection{Stacked Dense Neck}\label{network:neck}
The Feature Pyramid Network (FPN) is a fundamental component in object detection frameworks, serving as an adaptive module that integrates and improves hierarchical features. It acts like a neck that connects the backbone and the subsequent detection heads. The original FPN design~\citep{lin2017feature} transfers multi-level semantic information from the backbone through a top-down pathway and lateral connections, creating a simple and straightforward path for knowledge integration. Subsequent designs~\citep{liu2018path,tan2020efficientdet,pang2019libra,ghiasi2019fpn} have introduced cross-scale connections to reinforce visual representations with semantically important information and low-level details.

We employ a stacked, densely connected FPN, namely NAS-FPN~\citep{ghiasi2019fpn}, as the neck of our object detector for the following four reasons.
\begin{itemize}
\item As universal object detection is to detect hundreds of object categories from various datasets, the impressive ability of NAS-FPN to generate robust representations meets the challenges of million-scale multi-domain detection.
\item As we freeze the backbone, the remaining detector components require higher model capacity (described in Section~\ref{network:detector}), while stacked NAS-FPN offers excellent flexibility in constructing rich neck architecture.
\item The released SEER models are trained on billions of uncurated web-scale images.
Inevitably, there is some domain gap between the upstream pre-training dataset and downstream detection datasets.
As we do not finetune the SEER models on the downstream data, we believe the early NAS-FPN blocks can act as domain adaptors to align the domain gap.
\item Last but not least, we observe that multi-level side-outputs of SEER models have very different characteristics. Some shallow side-outputs are dense, while the deeper ones are generally sparse and weak.
The rich connections of NAS-FPN offer more possible ways for better feature integration.
\end{itemize}

\subsubsection{Adaptive RPN}\label{network:rpn}
In the context of multi-domain object detection, objects belonging to the same category can exhibit different characteristics across different domains. For example, a \emph{person} in an autonomous driving dataset such as MVD is typically much smaller in the high-resolution street scenes, while a \emph{person} in COCO images is usually much larger. This variation in object size highlights the need for an adaptive region proposal network (RPN) to generate high-quality proposals that can handle the diverse object sizes in each domain. The Cascade RPN~\citep{vu2019cascade} overcomes the limitations of traditional RPNs, which rely on heuristically determining appropriate scales and aspect ratios for pre-defined anchors. Additionally, having too many pre-defined anchors can slow down the training process. By incorporating the Cascade RPN into our network, we are able to improve the quality of proposals and increase the overall model capacity, providing the best of both worlds.

\subsection{Cross-dataset Model Training}
\subsubsection{Label Space Unification Across Multiple Datasets}\label{uls}
The goal of this section is to outline the creation of a unified label space for three datasets, addressing the issues of label duplication and semantic hierarchy across the datasets. The RVC organizers have provided a manually-crafted taxonomy\footnote{\url{https://github.com/ozendelait/rvc\_devkit/blob/master/objdet/obj\_det\_mapping.csv}} as a starting point. This taxonomy maps each category from COCO or MVD to a single category in the RVC official label space, as well as each leaf-node category from OID. However, the non-leaf categories from OID are not included in this label space. To complete the label space, we modify the official taxonomy by simply adding all of the non-leaf categories from OID, excluding the root entry. This results in a unified label space with a cardinality of 540.

The OID has a semantic hierarchy where the superclasses, or non-leaf categories, are considered to be more general than other classes. However, this leads to inconsistencies in granularity and results in issues such as label duplication and problems with the semantic hierarchy. For example, the \emph{person} (/m/01g317) superclass in OID and the \emph{person} class in COCO are semantically the same, but are treated as separate categories. For another example, the \emph{cow} class in COCO is semantically a child of the \emph{animal} (/m/0jbk) superclass in OID, but OID's hierarchy does not reflect this parent-child relationship. This overlap in taxonomy can negatively impact the performance of universal object detection. To address these obstacles, we propose a unified hierarchical taxonomy and implement a hierarchy-aware loss suppression method, which will be explained in Section~\ref{taxnomy} and \ref{weight}, respectively.

\subsubsection{Multi-label with Hierarchical Taxonomy Completion}\label{taxnomy}
To address the semantic hierarchy challenges in OID, we introduce a method of completing the hierarchical taxonomy by incorporating categories from the RVC official taxonomy. The resolution of the remaining cross-dataset semantic hierarchy issues is presented in Section~\ref{weight}.

We convert the one-hot category labels to multi-class labels by considering all parent categories as positives for OID images. This is similar to UniDet~\citep{zhou2022simple} but with the added consideration of the semantic hierarchies of COCO and MVD using the OID semantic hierarchy. For each annotated box that has been merged with an OID leaf-node category according to the RVC official taxonomy, we treat it as its OID equivalent. For instance, if the COCO \emph{banana} and the OID \emph{banana} (/m/09qck) have been merged into a single mutual category, a bounding box annotated as \emph{banana} from COCO would receive a positive label for the \emph{fruit} category since \emph{banana} belongs to the \emph{fruit} superclass according to the OID semantic hierarchy. We employ a multi-label classifier in the detection heads and use \emph{sigmoid} activation functions to obtain class confidence scores for each bounding box.

It is important to note that this hierarchical taxonomy completion is not a complete solution. There are several annotated objects from COCO and MVD that do not match any OID leaf-node category but are semantically associated with a certain superclass from OID. Instead of activating the corresponding parent categories, we handle these semantic hierarchies through an intricate adaptation in the loss function, which is discussed in the following section.

\begin{figure}[!t]
  \centering
  \includegraphics[width=2.8in]{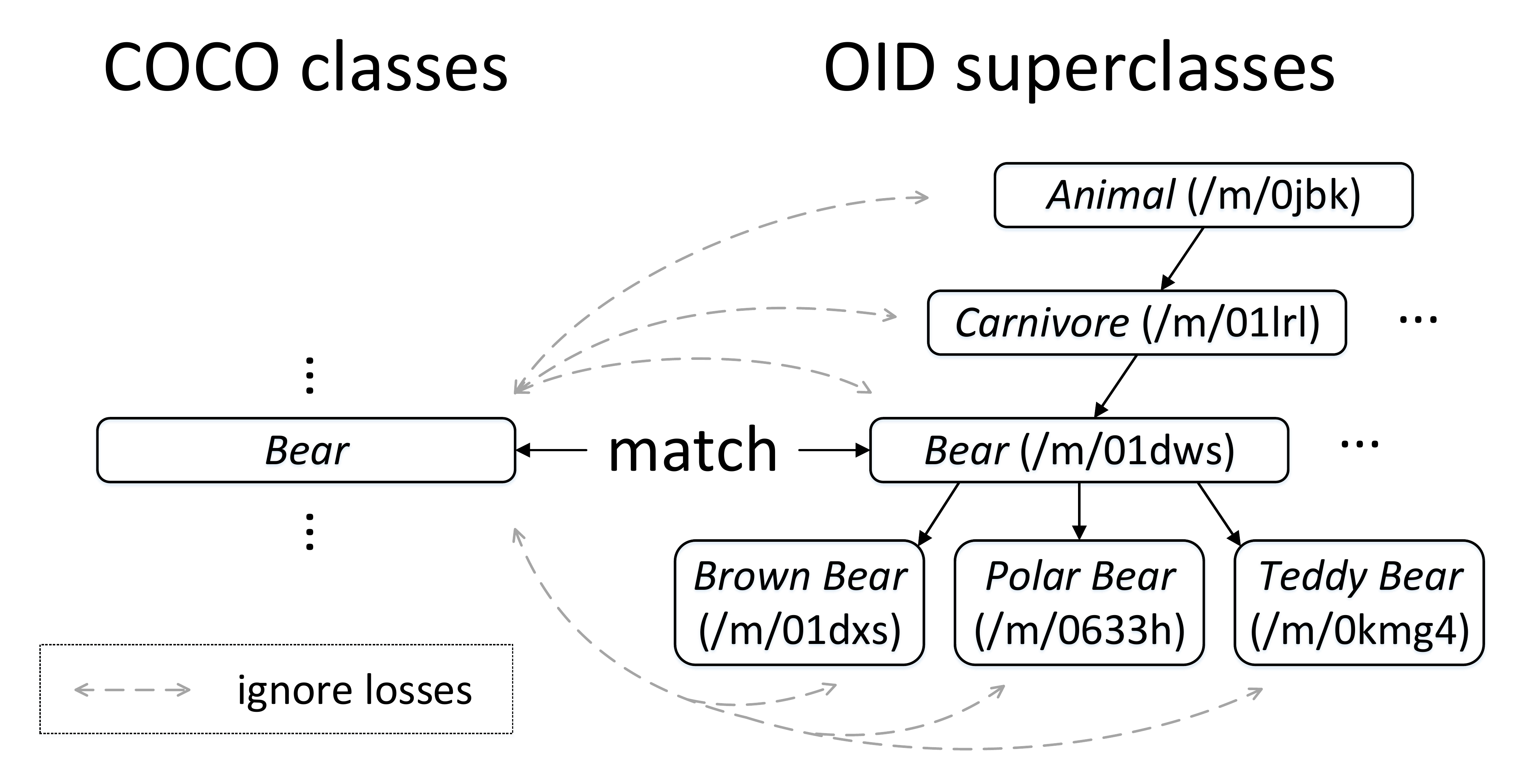}
  \caption{An example shows the loss suppression for semantically label duplication between COCO and OID.}\label{fig:loss1}
\end{figure}

\subsubsection{Hierarchy-aware Cross-dataset Loss Suppression}\label{weight}
To address both label duplication and unsolved semantic hierarchies described in Sections~\ref{uls} and \ref{taxnomy}, we propose a loss adaptation strategy called Hierarchy-Aware Cross-Dataset Loss Suppression (HCLS). This strategy is based on the semantic hierarchy of OID and suppresses losses over categories involved in label duplication and semantic hierarchy between OID and COCO/MVD in the box classification branches. More specifically,

\begin{itemize}
\item For each category from OID, HCLS ignores the losses over all its child categories, as a common practice for hierarchical taxonomy~\citep{kuznetsova2020open}.
\item For each category from COCO / MVD, which is not merged with any OID leaf-node category in the RVC official taxonomy, HCLS searches all the superclasses from OID and performs one of the following adaptations to the loss:
\subitem(a)~[Label duplication] Suppose this category matches one of the superclasses from OID in semantics. In this case, HCLS ignores the loss between its OID equivalent and itself, in addition to the losses between its OID equivalent’s parents/children and itself.
\subitem(b)~[Cross-dataset semantic hierarchy] Suppose this category belongs to one of the superclasses from OID in semantics. 
HCLS ignores the losses between all its parent categories and itself.
\subitem(c)~[Neither] Suppose this category is independent of any superclass of OID.
HCLS does nothing about loss adaptation.
In other words, we equally calculate losses over all the categories in the unified label space in loss functions.
\end{itemize}

\begin{figure}[!t]
  \centering
  \includegraphics[width=2.8in]{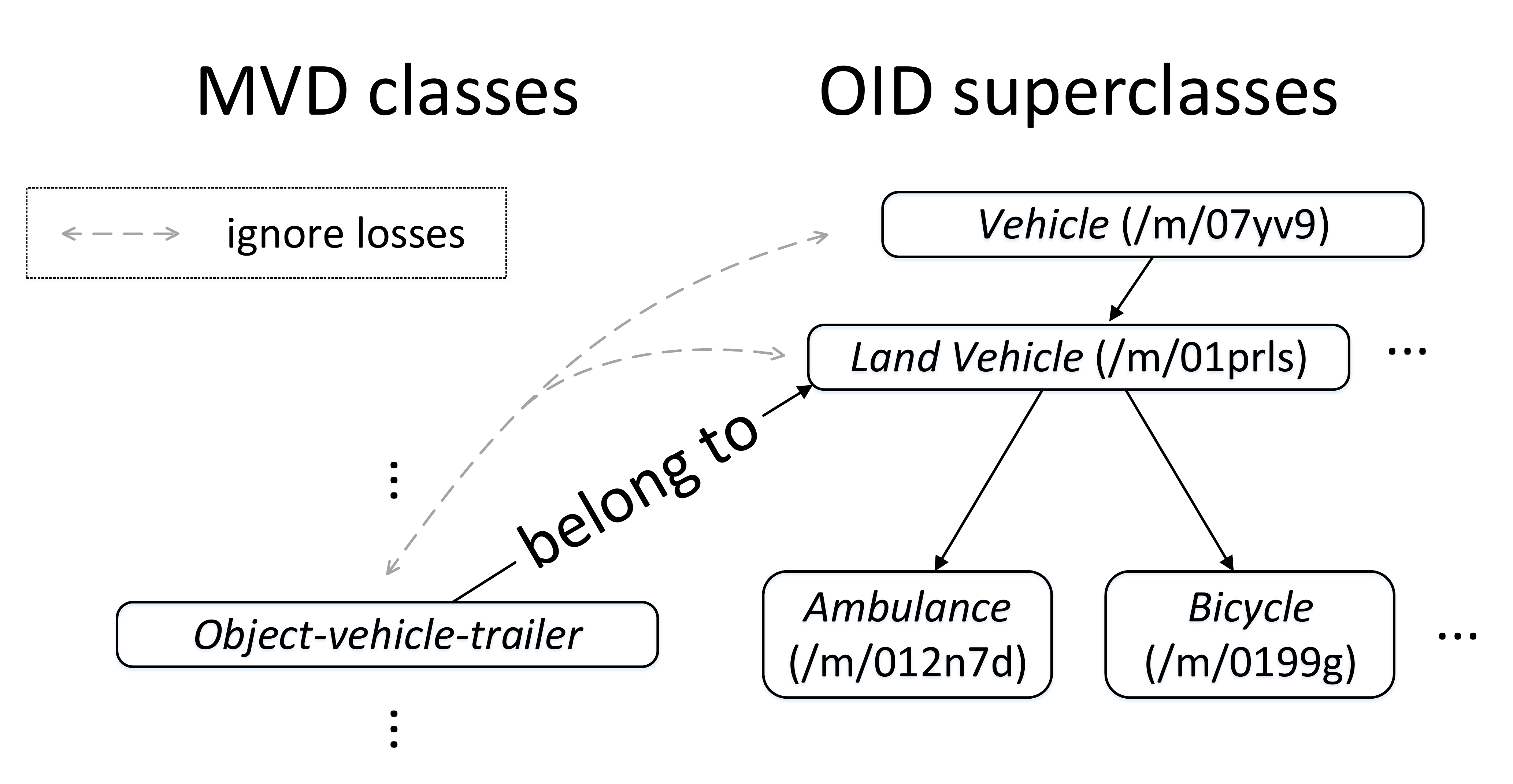}
  \caption{An example shows the loss suppression for semantic hierarchy between MVD and OID.}\label{fig:loss2}
\end{figure}

In Fig.~\ref{fig:loss1} and Fig.~\ref{fig:loss2}, two examples illustrate the loss adaptation process: (a) and (b). Note that we do \textbf{not} perform any tedious category merging but rather handle label duplication at the loss level. According to the RVC official taxonomy, there are less than 50 independent categories, so we manually search for cross-dataset label duplication and semantic hierarchy. For further details, please refer to Table~\ref{tab:loss1}, which lists the processed semantically duplicate categories, and Table~\ref{tab:loss2}, which lists the processed semantic hierarchies across datasets.

\begin{table}[!bp]
\small
\begin{center}
\begin{tabular}{p{90pt}p{100pt}}
\toprule
\multicolumn{2}{c}{Two categories match in semantics} \\
\midrule
~~COCO classes & OID superclasses  \\
\midrule
~~\emph{sports ball} & \emph{ball} (/m/018xm) \\
~~\emph{bear} & \emph{bear} (/m/01dws) \\
~~\emph{bed} & \emph{bed} (/m/03ssj5) \\
~~\emph{bird} & \emph{bird} (/m/015p6) \\
~~\emph{boat} & \emph{boat} (/m/019jd) \\
~~\emph{car} & \emph{car} (/m/0k4j) \\
~~\emph{clock} & \emph{clock} (/m/01x3z) \\
~~\emph{person} & \emph{person} (/m/01g317) \\
\midrule
~~MVD classes & OID superclasses  \\
\midrule
~~\emph{object--vehicle--car} & \emph{car} (/m/0k4j) \\
~~\emph{human--person} & \emph{person} (/m/01g317) \\
\bottomrule
\end{tabular}
\smallskip\smallskip
\caption{The duplicated category names between OID superclasses and COCO / MVD classes in semantics.}\label{tab:loss1}
\end{center}
\end{table}

\subsubsection{Overall Formulation}\label{loss}
The overall loss function can be formulated as the weighted sum of the RPN loss and the detector head loss, described as follows,
\begin{equation}\label{eq1}
L = \lambda \cdot L_{rpn} + L_{head}
\end{equation}
where $\lambda$ is the weight factor set to 0.7, while $L_{rpn}$ represents the RPN loss~(\ref{eq2}) and $L_{head}$ represents the detector head loss~(\ref{eq3}). 
In the detector head loss $L_{head}$, the classification loss $L_{cls}$ is given in~(\ref{eq4}). 

In formulas~(\ref{eq2}), (\ref{eq3}), and (\ref{eq4}), the symbols $p$, $q$, $r$, and $x$ denote the respective outputs for RPN regression branch, RPN classification branch, the detector head regression branch, and the detector head classification branch, while the $y$ denotes the corresponding ground truth.
$N$ is the number of samples in each mini-batch.
$C$ is the number of categories including \emph{background} in the unified label space.
$S_{rpn}$ represents the number of stages of the Cascade RPN while $S_{head}$ represents the number of stages of Cascade R-CNN.
We set $S_{rpn}$ to 2 and $S_{head}$ to 3 for a performance-cost trade-off reason.
$IoU$ represents the IoU loss~\citep{yu2016unitbox}, $BCE$ represents the binary cross entropy loss~\citep{girshick2015fast}, and $SmoothL_1$ represents the smooth L1 loss~\citep{girshick2015fast}.
$\mathds{1}_{\mathds{A}}(x)$ represents the indicator function in which the result turns out 1 when $x \in \mathds{A}$.
Specifically, $\mathds{D}(y)$ denotes the union of the categories involved in HCLS for class $y$, as described in Section~\ref{weight}.
$\mathds{P}(y)$ denotes the union of the parent categories of class $y$ as described in Section~\ref{taxnomy}.
To achieve a balanced weighting of each component of the loss, we empirically set the weights $\alpha$, $\beta$, and $\gamma$ to 10.0, 1.0, and 1.5, respectively.

\begin{figure*}[!t]
\centering
\begin{align}\label{eq2}
L_{rpn} = \frac{1}{N} \sum_{i=0}^N \sum_{s=0}^{S_{rpn}} (\alpha \cdot (1 - {IoU}(p_{s}^{i},~y_{rloc}^{i})) + {BCE}(q_{s}^{i},~y_{rcls}^{i}))
\end{align}
\vspace{-10pt}
\begin{align}\label{eq3}
L_{head} = \frac{1}{N} \sum_{i=0}^N \sum_{s=0}^{S_{head}} (\beta \cdot {SmoothL_1}(r_{s}^{i},~y_{loc}^{i}) + \frac{\gamma}{C} \cdot \sum_{c=0}^{C} L_{cls}^c)
\end{align}
\vspace{-5pt}
\begin{align}\label{eq4}
L_{cls}^c = (1 - \mathds{1}_{\mathds{D}(y_{cls}^i)}(c)) \cdot {BCE}(x_{s}^{ic},~\mathds{1}_{\mathds{P}(y_{cls}^i)} (c))
\end{align}
\end{figure*}

\begin{table}[!bp]
\small
\begin{center}
\begin{tabular}{p{90pt}p{100pt}}
\toprule
\multicolumn{2}{c}{The left as a descendant of the right in semantics} \\
\midrule
~COCO classes & OID superclasses  \\
\midrule
~\emph{cow} & animal (/m/0jbk) \\
\midrule
~MVD classes & OID superclasses  \\
\midrule
~\emph{animal--ground} & \emph{animal} (/m/0jbk) \\
~\emph{--animal} & \\
~\emph{object--vehicle} & \emph{land vehicle} (/m/01prls) \\
~\emph{--caravan} & \\
~\emph{object--vehicle--other} & \emph{land vehicle} (/m/01prls) \\
~\emph{--vehicle} & \\
~\emph{object--vehicle--trailer} & \emph{land vehicle} (/m/01prls) \\
~\emph{object--vehicle} & \emph{land vehicle} (/m/01prls) \\
~\emph{--wheeled--slow} & \\
~\emph{object--support--traffic} & \emph{traffic sign} (/m/01mqdt) \\
~\emph{--sign--frame} & \\
~\emph{object--traffic--sign} & \emph{traffic sign} (/m/01mqdt) \\
~\emph{--back} & \\
~\emph{object--traffic--sign} & \emph{traffic sign} (/m/01mqdt) \\
~\emph{--front} & \\
\bottomrule
\end{tabular}
\smallskip\smallskip
\caption{The parent-child category names between OID superclasses and COCO / MVD classes in semantics.}\label{tab:loss2}
\end{center}
\end{table}

\section{Experiments}
\subsection{Datasets}

In Table~\ref{tab:datsets}, we present a concise overview of the three datasets employed in our experiments. The COCO dataset~\citep{lin2014microsoft} comprises everyday images featuring objects and humans, with annotations for 80 common object categories. The MVD dataset~\citep{neuhold2017mapillary} (version 1.2) provides high-resolution street-scene imagery and includes 37 object categories. In contrast to COCO and MVD, the OID dataset~\citep{kuznetsova2020open} stands out with its annotated semantic hierarchy. It features diverse images, often containing multiple objects and complex scenes, and exhibits a long-tailed distribution of annotated classes. For our experiments, we use the training set from the OID dataset, which includes 500 object categories, adhering to the standard practice of the Open Images Challenge 2019.

\begin{table}[!tbp]
\centering\small
\begin{tabular}{p{36pt}|p{70pt}<{\centering}p{30pt}<{\centering}p{30pt}<{\centering}}
\hline
\rule{0pt}{10pt}~Dataset & Domain & \# Cats & \# Imgs \\
\hline
\rule{0pt}{10pt}~COCO & Internet images & 80 & 118k \\
\rule{0pt}{10pt}~MVD & Street scenes & 37 & 1.8k \\
\rule{0pt}{10pt}~OID & Internet images & 500 & 1.8M \\
\hline
\end{tabular}
\smallskip\smallskip
\caption{The training datasets used in the experiments, which are provided by organizers of RVC 2022.}\label{tab:datsets}
\end{table}

\subsection{Implementation Details}\label{implement}
For our experiments, we implement our proposed method using the mmdetection codebase~\citep{chen2019mmdetection}. We use the frozen SEER-RegNet32gf and SEER-RegNet256gf as the backbone models, which offer the resource-efficient training. To ensure synchronized computation across all GPU workers, we replace traditional BatchNorm (BN) with synchronized BatchNorm (SyncBN). Hyper-parameters for the NAS-FPN, Cascade RPN, and Cascade R-CNN components are maintained at their default settings, unless otherwise specified.


We apply standard data augmentation techniques, including random flipping and random scaling of the short edge of the image within the range of [480,~960]. For optimization, we employ the Stochastic Gradient Descent (SGD) optimizer with a base learning rate of 0.01, weight decay set to 0.0001, and a batch size of 16. To handle imbalanced class distribution and size disparities across the three datasets, we use class-aware sampling and dataset-wise re-sampling. The re-sampling ratio is configured as 1:~4:~8 for OID, COCO, and MVD, respectively. Model training is conducted on 8 NVIDIA 3090 / A100 GPUs, with the use of mixed-precision techniques~\citep{micikevicius2018mixed} for accelerated training. The models are trained for 1.15 million iterations. The training process incorporates a linear warm-up of the base learning rate in the initial 4k iterations, followed by a reduction by a factor of 10 at 850k and 1.0M iterations. During the inference stage, images are resized with the short edge set to 800 and the long edge constrained to 1333, maintaining the aspect ratio, unless otherwise specified.

\begin{table*}[!t]
\centering\small
\begin{tabular}{lp{22pt}<{\centering}p{22pt}<{\centering}p{22pt}<{\centering}p{22pt}<{\centering}p{22pt}<{\centering}p{22pt}<{\centering}p{22pt}<{\centering}p{22pt}<{\centering}p{34.5pt}<{\centering}p{34.5pt}<{\centering}}
\cmidrule(r){1-11}
\multirow{2}{*}{Methods} & \multicolumn{2}{c}{COCO~\emph{test}} & \multicolumn{2}{c}{COCO~\emph{val}} & \multicolumn{2}{c}{MVD~\emph{test}} & \multicolumn{2}{c}{MVD~\emph{val}} & OID~\emph{test} & OID~\emph{val} \\
\cmidrule(r){2-11}
& mAP & AP50 & mAP & AP50 & mAP & AP50 & mAP & AP50 & AP50 & AP50 \\
\cmidrule(r){1-1}\cmidrule(r){2-3}\cmidrule(r){4-5}\cmidrule(r){6-7}\cmidrule(r){8-9} \cmidrule(r){10-11}
MD\_RVC (1) & 59.0 & 76.0 & - & - & 32.8 & 49.3 & - & - & 62.1 & - \\
IFFF\_RVC (2) & 50.0 & 69.0 & 50.0 & 69.1 & 25.3 & 39.0 & 24.2 & 38.6 & 59.9 & 69.1 \\
USTC\_RVC (3) & 50.0 & 68.0 & - & - & 25.1 & 37.9 & - & - & 47.8 & - \\
CBS\_RVC (4) & 48.0 & 66.0 & - & - & 13.5 & 21.5 & - & - & 47.7 & - \\
TSDREF\_RVC (5) & 19.0 & 31.0 & - & - & 6.6 & 11.6 & - & - & 40.2 & - \\
\cmidrule(r){1-11}
Large-UniDet [S] & - & - & 48.8 & 66.2 & - & - & 25.9 & 39.4 & - & 68.5 \\
Large-UniDet [L] & - & - & \textbf{51.9} & \textbf{70.0} & - & - & \textbf{27.7} & \textbf{42.2} & - & \textbf{69.8} \\
\cmidrule(r){1-11}
Large-UniDet [S]$^{\dag}$ & - & - & 52.0 & 70.4 & - & - & 32.0 & 47.8 & - & 69.2 \\
Large-UniDet [L]$^{\dag}$ & - & - & \textbf{53.5} & \textbf{71.8} & - & - & \textbf{33.2} & \textbf{49.4} & - & \textbf{70.5} \\
\cmidrule(r){1-11}
\end{tabular}
\smallskip
\caption{Comparisons on five RVC submissions on three datasets, the numbers in brackets denote the achieved places in the challenge. The last four rows report our updated results in this paper, we only provide the accuracy on validation sets as the RVC test server is off after RVC deadline. Large-UniDet [S] indicates our method based on SEER-RegNet32gf while Large-UniDet [L] indicates the one based on SEER-RegNet256gf.}\label{tab:result}
\end{table*}

\subsection{Main Results}

In Table~\ref{tab:result}, we provide a summary of comparisons between the RVC final submissions and our latest results. The top-ranking method, MD\_RVC, utilizes a large transformer-based object detector~\citep{carion2020end} combined with an acceleration training strategy that progressively increases the input size. In contrast, our method, named Large-UniDet, takes a quite different approach. We focus on developing an efficient training formulation to save both computation and memory resources. Our goal is to generate robust multi-domain object detection predictions by leveraging the capabilities of large pre-trained vision models.


Compared to our RVC submission, IFFF\_RVC (detailed in Appendix~\ref{simply}), the latest Large-UniDet not only retains its complexity but also incorporates further improvements through a training practice that adapts the model for high-resolution input data (detailed in Section~\ref{tune}). As evident from Table~\ref{tab:result}, with a lighter backbone (SEER-RegNet32gf), Large-UniDet achieves notable performance metrics: 48.8 mAP on COCO \emph{val} set, 66.2 AP50 on COCO \emph{val} set, 25.9 mAP on MVD \emph{val} set, 39.4 AP50 on MVD \emph{val} set, and 68.5 AP50 on OID \emph{val} set. When moving to the larger backbone (SEER-RegNet256gf), we consistently observe improved universal object detection performance, with a gain of +3.1 mAP and +3.8 AP50 on COCO, +1.8 mAP and +2.8 AP50 on MVD, and +1.3 AP50 on OID, which underscore the effectiveness of visual representations generated by larger vision models. Please note that we resized the short edges of MVD images to \textbf{2048} pixels, in alignment with our approach for the submission IFFF\_RVC. Additionally, we restricted the generation of predictions to a maximum of 300 per image, utilizing standard Non-Maximum Suppression (NMS) during testing to ensure a fair comparison with IFFF\_RVC on validation sets.

After completing universal object detection training, we proceed with dataset-specific individual finetuning using high-resolution training images, as elaborated in Section~\ref{tune}. This practice leads to further performance improvements on all three datasets, with particularly significant gains observed for MVD, which exhibits distinct characteristics compared to the other two datasets. Detailed AP numbers reflecting these improvements are presented in Table~\ref{tab:result} and are denoted as Large-UniDet with superscript~$^\dag$.

\subsection{Ablation Study}
\subsubsection{Detector Components Analysis}\label{exp:det_components}
We perform an ablation analysis using SEER-RegNet32gf as the backbone. The results of this analysis, focusing on accuracy-cost comparisons for different detector configurations, are presented in Table~\ref{tab:det_components_coco} for COCO and Table~\ref{tab:det_components_mvd} for MVD. When employing a frozen backbone, Cascade R-CNN exhibits a significant performance improvement over the baseline Faster R-CNN. This improvement is substantial, raising mAP to 39.9 on COCO and 15.2 on MVD, with only a slight increase in training cost (+1 hour for COCO and +0.2 hours for MVD).


When integrating high-capacity necks into the Cascade R-CNN, we observe that higher accuracy is achievable, but this comes at the expense of increased computation burden. We compare three commonly used FPNs in Tables~\ref{tab:det_components_coco} and \ref{tab:det_components_mvd}. As we can see, PAFPN~\citep{liu2018path} yields a modest improvement of 0.9 and 1.5 points on COCO and MVD, respectively. BiFPN~\citep{tan2020efficientdet} offers a slightly greater improvement, with at most 1.6 and 1.9 points on COCO and MVD, respectively. However, NAS-FPN surpasses them all, delivering remarkable gains of 5.8 and 4.2 points on COCO and MVD. Additionally, Tables~\ref{tab:det_components_coco} and \ref{tab:det_components_mvd} illustrate that as the number of stacked neck blocks increases, so does the computational cost. However, the rate of performance improvement gradually decelerates, and in some cases, such as the stacked BiFPN, performance even degrades. Hence, we opt for seven stacked NAS-FPN blocks for our detector, striking a favorable balance between accuracy and computational cost.


In addition, as observed, Cascade RPN consistently improves performance, contributing at least a +1.0 mAP gain on COCO and a +0.2 mAP gain on MVD, regardless of the neck configuration. This improvement is particularly notable when coupled with NAS-FPN, and it comes without a significant increase in extra computational cost.

\begin{table}[!tbp]
\centering\small
\begin{tabular}{lll}
\toprule
Model & mAP & Time (h) \\
\midrule
Faster R-CNN & 29.0 & 12 \\
\midrule
Cascade R-CNN & 39.9 & 13 \\
\rule{0pt}{12pt}~~~+ PAFPN & 40.8 \textcolor{cyan}{(+0.9)} & 16 \textcolor{red}{(+3)} \\
~~~~~~+ Cascade RPN & 42.0 \textcolor{cyan}{(+2.1)} & 19 \textcolor{red}{(+6)} \\
\rule{0pt}{12pt}~~~+ BiFPN \footnotesize{$(\times1)$} & 40.5 \textcolor{cyan}{(+0.6)} & 16 \textcolor{red}{(+3)} \\
~~~+ BiFPN \footnotesize{$(\times3)$} & 41.5 \textcolor{cyan}{(+1.6)} & 18 \textcolor{red}{(+5)} \\
~~~+ BiFPN \footnotesize{$(\times5)$} & 41.3 \textcolor{cyan}{(+1.4)} & 20 \textcolor{red}{(+7)} \\
~~~+ BiFPN \footnotesize{$(\times7)$} & 41.1 \textcolor{cyan}{(+1.2)} & 22 \textcolor{red}{(+9)} \\
~~~~~~+ Cascade RPN & 42.1 \textcolor{cyan}{(+2.2)} & 25 \textcolor{red}{(+12)} \\
\rule{0pt}{12pt}~~~+ NAS-FPN \footnotesize{$(\times1)$} & 41.6 \textcolor{cyan}{(+1.7)} & 16 \textcolor{red}{(+3)} \\
~~~+ NAS-FPN \footnotesize{$(\times3)$} & 44.2 \textcolor{cyan}{(+4.3)} & 18 \textcolor{red}{(+5)} \\
~~~+ NAS-FPN \footnotesize{$(\times5)$} & 45.3 \textcolor{cyan}{(+5.4)} & 20 \textcolor{red}{(+7)} \\
~~~+ NAS-FPN \footnotesize{$(\times7)$} & 45.7 \textcolor{cyan}{(+5.8)} & 22 \textcolor{red}{(+9)} \\
~~~~~~+ Cascade RPN & 47.6 \textcolor{cyan}{(+7.7)} & 24 \textcolor{red}{(+11)} \\
\bottomrule
\end{tabular}
\smallskip\smallskip
\caption{Ablation analysis of detector components on COCO \emph{val} set. The models are trained for 12 epochs on 8 NVIDIA 3090 GPUs, with a base learning rate 0.01 which is divided by 10 after 8 and 11 epochs.}\label{tab:det_components_coco}
\end{table}

\begin{table}[!tbp]
\centering\small
\begin{tabular}{lll}
\toprule
Model & mAP & Time (h) \\
\midrule
Faster R-CNN & 12.8 & 1.8 \\
\midrule
Cascade R-CNN & 15.2 & 2.0 \\
\rule{0pt}{12pt}~~~+ PAFPN & 16.7 \textcolor{cyan}{(+1.5)} & 2.5 \textcolor{red}{(+0.5)} \\
~~~~~~+ Cascade RPN & 17.3 \textcolor{cyan}{(+2.1)} & 3.5 \textcolor{red}{(+1.5)} \\
\rule{0pt}{12pt}~~~+ BiFPN \footnotesize{$(\times1)$} & 17.0 \textcolor{cyan}{(+1.8)} & 2.5 \textcolor{red}{(+0.5)} \\
~~~+ BiFPN \footnotesize{$(\times3)$} & 17.1 \textcolor{cyan}{(+1.9)} & 2.8 \textcolor{red}{(+0.8)} \\
~~~+ BiFPN \footnotesize{$(\times5)$} & 16.8 \textcolor{cyan}{(+1.6)} & 3.1 \textcolor{red}{(+1.1)} \\
~~~+ BiFPN \footnotesize{$(\times7)$} & 16.5 \textcolor{cyan}{(+1.3)} & 3.4 \textcolor{red}{(+1.4)} \\
~~~~~~+ Cascade RPN & 16.7 \textcolor{cyan}{(+1.5)} & 4.3 \textcolor{red}{(+2.3)} \\
\rule{0pt}{12pt}~~~+ NAS-FPN \footnotesize{$(\times1)$} & 17.3 \textcolor{cyan}{(+2.1)} & 2.8 \textcolor{red}{(+0.8)} \\
~~~+ NAS-FPN \footnotesize{$(\times3)$} & 18.3 \textcolor{cyan}{(+3.1)} & 3.0 \textcolor{red}{(+1.0)} \\
~~~+ NAS-FPN \footnotesize{$(\times5)$} & 18.6 \textcolor{cyan}{(+3.4)} & 3.2 \textcolor{red}{(+1.2)} \\
~~~+ NAS-FPN \footnotesize{$(\times7)$} & 19.4 \textcolor{cyan}{(+4.2)} & 3.5 \textcolor{red}{(+1.5)} \\
~~~~~~+ Cascade RPN & 20.2 \textcolor{cyan}{(+5.0)} & 4.2 \textcolor{red}{(+2.2)} \\
\bottomrule
\end{tabular}
\smallskip\smallskip
\caption{Ablation analysis of detector components on MVD \emph{val} set. For fast convergence, we initialize the models with the counterparts in Table~\ref{tab:det_components_coco}, and train them for 12 epochs on 8 NVIDIA 3090 GPUs, with a base learning rate 0.01 which is divided by 10 after 8 and 11 epochs.}\label{tab:det_components_mvd}
\end{table}

\subsubsection{Hierarchical Loss Strategies}\label{losses}
To demonstrate the effectiveness of our hierarchy-aware cross-dataset loss suppression (HCLS), we compare a number of hierarchical loss strategies in Table~\ref{tab:label1} and \ref{tab:label2}. The used object detector in Table~\ref{tab:label1} is Cascade R-CNN based on SEER-RegNet32gf, while the used object detector in Table~\ref{tab:label2} is Cascade R-CNN based on SEER-RegNet32gf with NAS-FPN and Cascade RPN. For a quick evaluation, the five models in the tables are trained for only 420k iterations, with a base learning rate of 0.01, which is reduced by a factor of 0.1 at 280k iterations.

\begin{itemize}
\item \emph{Baseline} refers to a situation where the semantic hierarchy is not taken into consideration, and each annotated bounding box is assigned a single positive class label. As a result, there is no loss adaptation applied.

\item \emph{Naive loss suppression} denotes that the loss calculation for the classification task takes the semantic hierarchy of OID into account by ignoring the losses for the children and parent categories. This approach incorporates the semantic hierarchy by removing the impact of relationships between parent and child categories, but also leads to an evident loss of positive samples for the superclasses in OID, resulting in lower performance on OID.

\item \emph{Unified hierarchy} takes into account all parent-child relationships across datasets by considering the cross-dataset label duplication presented in Table~\ref{tab:loss1}, cross-dataset semantic hierarchy presented in Table~\ref{tab:loss2}, and the original semantic hierarchy of OID for each category in the unified label space. It considers all parents and semantic equivalents as positive and eliminates the losses over all child categories, thereby increasing the training set for superclasses with more positive samples, resulting in a significant improvement in performance on OID (+3.1 AP50 in Table~\ref{tab:label1} and +1.8 AP50 in Table~\ref{tab:label2}). However, this approach may negatively impact the performance on COCO and MVD, as categories from different datasets may match based on language cues, but still be semantically different. For example, the \emph{bear} category in COCO encompasses a wide range of carnivorous mammals of the Ursidae family, while its equivalent \emph{bear} (/m/01dws) in OID includes not only these conventional bears but also \emph{teddy bears} (/m/0kmg4), leading to taxonomy inconsistencies. This was observed in the severe performance decline of the \emph{bear} category in COCO, with an AP of 41.5 using the \emph{Unified hierarchy}, compared to APs over 65.1 for other categories in Table~\ref{tab:label1}, with the best AP of 67.9 achieved by the HCLS.

\item \emph{OID hierarchy} only takes into account the semantic hierarchy of OID. It does not consider the relationships between categories from different datasets. This is a common approach when working with OID~\citep{zhou2022simple}, but it means that cross-dataset relationships are not incorporated into the loss adaptation.

\item Our loss strategy, denoted as \emph{OID hierarchy} + \emph{HCLS}, cleverly incorporates the semantic hierarchy of OID while addressing label duplication and accounting for the overarching semantic hierarchy across all three datasets. This adaptive loss computation approach yields remarkable performance, as evidenced by achieving the highest accuracy on OID, with a significant +3.5 AP50 improvement according to Table~\ref{tab:label1} and +2.7 AP50 according to Table~\ref{tab:label2}. Furthermore, this strategy offers a modest enhancement on COCO and maintains performance on par with the \emph{Baseline} accuracy on MVD. Intriguingly, \emph{Naive loss suppression} performs comparably, and at times, even slightly better than our designed loss strategy on COCO and MVD. However, considering the substantial performance improvement observed on OID, we believe our method remains notably superior.
\end{itemize}

\begin{table}[!t]
\centering\small
\begin{tabular}{lccc}
\toprule
Loss Strategy & COCO & MVD & OID  \\
\midrule
Baseline & 36.2 & 14.4 & 61.7 \\
Naive loss suppression & 36.4 & \textbf{14.5} & 62.7 \\
Unified hierarchy & 36.3 & 14.0 & 64.8 \\
OID hierarchy & 36.4 & 14.0 & 64.8 \\
~~~+ HCLS & \textbf{37.1} & 14.3 & \textbf{65.2} \\
\bottomrule
\end{tabular}
\smallskip\smallskip
\caption{Comparison on loss strategies. The metrics for COCO and MVD are mAP, and for OID, it is AP50. The best results are highlighted in \textbf{bold font}.}\label{tab:label1}
\end{table}

\begin{table}[!t]
\centering\small
\begin{tabular}{lccc}
\toprule
Loss Strategy & COCO & MVD & OID  \\
\midrule
Baseline & 44.1 & 17.2 & 65.3 \\
Naive loss suppression & \textbf{44.3} & \textbf{17.8} & 66.2\\
Unified hierarchy & 43.1 & 17.1 & 67.1 \\
OID hierarchy & 43.5 & 16.2 & 67.6 \\
~~~+ HCLS & 44.2 & 17.2 & \textbf{68.0} \\
\bottomrule
\end{tabular}
\smallskip\smallskip
\caption{Comparison on loss strategies. The metrics for COCO and MVD are mAP, and for OID, it is AP50. The best results are highlighted in \textbf{bold font}.}\label{tab:label2}
\end{table}

\begin{table*}[!t]
\centering\small
\begin{tabular}{p{35pt}<{\centering}p{50pt}<{\centering}p{45pt}<{\centering}p{62pt}p{62pt}p{30pt}<{\centering}p{30pt}<{\centering}p{45pt}}
\toprule
Dataset & Backbone~~~ & Strategy & ~~~~~mAP & ~~~~AP50 & epochs & Time & ~Memory \\
\midrule
\multirow{4}{*}{\rule{0pt}{12pt}COCO} & \multirow{4}{*}{\thead{\rule{0pt}{12pt}SEER-~~~\\RegNet32gf~~~}} & 
finetune & ~~~~~50.7 & ~~~~68.8 & 45 & 167 &~~~~16 \\
& & freeze & ~~~~~50.9~\textcolor{cyan}{(+0.2)} & ~~~~69.2~\textcolor{cyan}{(+0.4)} & 72 & 160 &~~~~10~\textcolor[rgb]{0.15,0.8,0.3}{(--6)} \\
\cmidrule(r){3-8}
& & finetune & ~~~~~49.9 & ~~~~68.3 & 64 & 238 &~~~~16 \\
& & freeze & ~~~~~51.4~\textcolor{cyan}{(+1.5)} & ~~~~69.8~\textcolor{cyan}{(+1.5)} & 108 & 240 &~~~~10~\textcolor[rgb]{0.15,0.8,0.3}{(--6)} \\
\midrule
\multirow{6}{*}{\rule{0pt}{24pt}MVD} & \multirow{4}{*}{\thead{\rule{0pt}{12pt}SEER-~~~\\RegNet32gf~~~}} & 
finetune & ~~~~~23.9 & ~~~~37.7 & 48 & 30 &~~~~16 \\
& & freeze & ~~~~~24.1~\textcolor{cyan}{(+0.2)} & ~~~~38.0~\textcolor{cyan}{(+0.3)} & 80 & 30 &~~~~10~\textcolor[rgb]{0.15,0.8,0.3}{(--6)} \\
\cmidrule(r){3-8}
& & finetune & ~~~~~24.4 & ~~~~38.2 & 100 & 61 &~~~~16 \\
& & freeze & ~~~~~24.8~\textcolor{cyan}{(+0.4)} & ~~~~39.1~\textcolor{cyan}{(+0.9)} & 160 & 60 &~~~~10~\textcolor[rgb]{0.15,0.8,0.3}{(--6)} \\
\cmidrule(r){3-8}
& \multirow{2}{*}{\thead{\rule{0pt}{2pt}SEER-~~\\RegNet256gf~~}} & 
finetune & ~~~~~25.8 & ~~~~40.4 & 60 & 70 &~~~~60 \\
& & freeze & ~~~~~26.0~\textcolor{cyan}{(+0.2)} & ~~~~40.9~\textcolor{cyan}{(+0.5)} & 120 & 72 &~~~~15~\textcolor[rgb]{0.15,0.8,0.3}{(--45)} \\
\bottomrule
\end{tabular}
\smallskip\smallskip
\caption{The object detectors are Cascade R-CNN enhanced with NAS-FPN ($\times7$) and Cascade RPN. The training time is measured in hours and the GPU memory consumption is measured in GB / image. The models are trained with a batch size of 16 on 16 NVIDIA 3090 / A100 GPUs. For comparison purpose, we convert the training time on different devices to the training time on 8 NVIDIA 3090 GPUs uniformly.}\label{tab:ft2}
\end{table*}

\subsubsection{Training Strategies}\label{training_strategies}
Table~\ref{tab:ft2} provides a comparison between two training approaches: finetuning the entire object detector, denoted as \emph{finetune}, and the approach used in this paper, where the backbone is frozen during training, denoted as \emph{freeze}. We evaluate these strategies by training models on either COCO or MVD, using both the lighter SEER backbone and the larger one. We assess their performance and GPU memory consumption.

\begin{itemize}
\item[-] \textbf{Performance}

The results of \emph{finetune} and \emph{freeze} are summarized in Table~\ref{tab:ft2}. It's evident that \emph{freeze} consistently enhances performance across datasets, particularly with an extended training schedule. In contrast, \emph{finetune} exhibits a decline in performance with a longer training schedule. This decline might be attributed to the backbone representations deviating from the original SEER visual representations and over-fitting on the smaller downstream dataset, thereby diminishing the object detector's performance.

\item[-] \textbf{GPU memory consumption}

As evident from Table~\ref{tab:ft2}, freezing the backbone during training imposes a substantially lower GPU memory demand compared to finetuning the entire object detector, including the backbone. Specifically, the SEER-RegNet32gf-based model requires 10 GB of GPU memory per image during training with the \emph{freeze} strategy, whereas the \emph{finetune} strategy demands 16 GB of GPU memory per image. Similarly, the SEER-RegNet256gf-based model requires 15 GB of GPU memory per image with the \emph{freeze} strategy, while the \emph{finetune} strategy necessitates a much higher 60 GB of GPU memory for each image. For this reason, freezing the backbone emerges as a more feasible option for training models on memory-constrained computational resources, such as NVIDIA 3090.
\end{itemize}

\subsubsection{Impact of Universal Paradigm}

To investigate the impact of the universal training paradigm, where the object detector learns from multi-domain datasets with a unified label spaces simultaneously, we conducted an ablation analysis.\footnote{The used object detectors are built upon Cascade R-CNN enhanced with NAS-FPN (×7) and Cascade RPN, utilizing SEER-RegNet32gf as the backbone. As detailed in Section~\ref{implement}, the universal object detector undergoes training for 1.15M iterations with a batch size of 16. Considering the re-sampled training set size of approximately 2.3M, the calculated training epochs amount to 8. Consequently, for Individual OD on OID, the training epochs are set to 8. On a comparable scale, for Individual OD on COCO, the object detector is trained for 12 epochs in line with the default configuration in the mmdetection codebase. Lastly, for Individual OD on MVD, the object detector is also trained for 12 epochs, initializing its network parameters with the object detector mentioned earlier for COCO.}
Table~\ref{tab:imp} compares the results achieved by the proposed universal object detector (Universal OD) with object detectors trained individually on their respective datasets (Individual OD).

As observed in Table~\ref{tab:imp}, the universal object detector consistently achieves superior results on all the three datasets. These findings suggest that universal object detection could potentially benefit from cross-domain annotations, which might help the detector in capturing more distinct visual concepts from one domain and applying them effectively in another.


\begin{table}[!t]
\centering\small
\begin{tabular}{p{70pt}<{\centering}p{30Pt}<{\centering}p{30Pt}<{\centering}p{30Pt}<{\centering}}
\toprule
Paradigm & COCO & MVD & OID  \\
\midrule
Universal OD & 48.8 & 21.1 & 68.5 \\
Individual OD & ~~~47.6~$\downarrow$ & ~~~20.2~$\downarrow$ & ~~~68.2~$\downarrow$ \\
\bottomrule
\end{tabular}
\smallskip\smallskip
\caption{Comparison of two training paradigms: universal object detection and individual object detection, using mAP for COCO and MVD, and AP50 for OID.}\label{tab:imp}
\end{table}

\subsubsection{Scaling Up and Finetuning}\label{tune}
\textbullet~\emph{Scaling up during the inference procedure}

Considering the larger size of MVD images and the prevalent occurrence of relatively small instances within them, a common scenario in computer vision, we adopted a strategic approach. As elaborated in Appendix~\ref{simply}, during the inference phase, we resized the short edges of MVD images to 2048 pixels. This scaling was implemented to enhance detection performance, particularly for the multitude of small objects present in high-resolution images. This scaling enhancement yielded an approximate 3-point increase in mAP on MVD within our RVC submission, yielding substantial benefits for the models showcased in this paper as well.

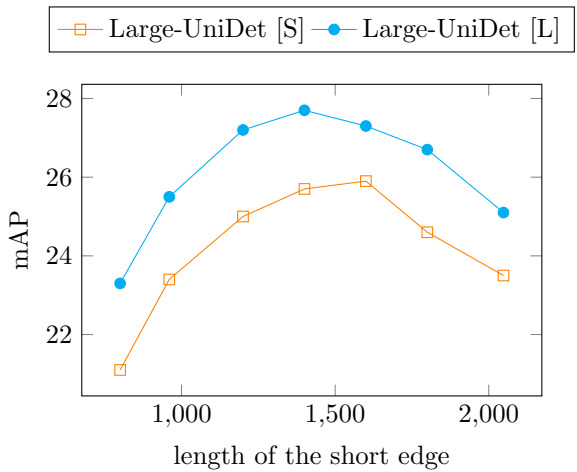
\begin{figure}[!t]
\centering
\begin{tikzpicture}
\begin{axis}[width=0.48\textwidth,height=2.25in,
xlabel=length of the short edge,ylabel=mAP,
legend style={at={(0.5,1.25)},anchor=north,legend columns=-1}]
\addplot+[sharp plot,mark=square,orange]
coordinates{
(800,21.1)
(960,23.4)
(1200,25.0)
(1400,25.7)
(1600,25.9)
(1800,24.6)
(2048,23.5)
};
\addplot[sharp plot,mark=*,cyan]
coordinates{
(800,23.3)
(960,25.5)
(1200,27.2)
(1400,27.7)
(1600,27.3)
(1800,26.7)
(2048,25.1)
};
\smallskip
\legend{Large-UniDet~[S],Large-UniDet~[L]}
\end{axis}
\end{tikzpicture}
\caption{The performance on MVD \emph{vs.} the lengths of the short edge of the testing images. We obtain the best mAP at 25.9 with the 1600-pixel resized short edge based on the lighter backbone (Large-UniDet~[S]), and obtain best mAP at 27.7 with the 1400-pixel resized short edge based on the larger one (Large-UniDet~[L]). The initial mAP and highest mAP of both two models are reported in Table~\ref{tab:up}.}\label{fig:mvd_len}
\end{figure}


As shown in Fig.~\ref{fig:mvd_len}, the models achieved optimal results when the testing image's short edge was resized to 1600 pixels (for Large-UniDet[S]) or 1400 pixels (for Large-UniDet[L]), as demonstrated in Table~\ref{tab:up} alongside the 800-pixel short-edge baselines. The effect of scaling is particularly pronounced in MVD performance for both the small and large models, resulting in a significant mAP increase of +4.8 and +4.4, respectively.


It is worth highlighting that increasing the testing image's size does not lead to performance enhancements always (\emph{e.g.} COCO or OID). This outcome arises from the fact that while scaling for inference aids in detecting smaller objects, it simultaneously compromises the performance for larger objects—a phenomenon previously documented in the literature~\citep{gao2018dynamic}. Bearing this in mind, we hold the belief that tailoring the object detector for high-resolution images bears the potential to enhance the detection of large objects while simultaneously retaining the benefits for smaller objects.

\begin{table}[!tbp]
\centering\small
\begin{tabular}{p{28pt}<{\centering}p{16pt}<{\centering}p{16pt}<{\centering}p{28pt}<{\centering}p{26pt}<{\centering}p{26pt}<{\centering}}
\toprule
Model & SU & FT & COCO & MVD & OID \\
\midrule
\multirow{3}{*}{[S]} & & & 48.8 & 21.1 & 68.5 \\
 & \checkmark & & - & 25.9 & - \\
 & \checkmark & \checkmark & \textbf{52.0} & \textbf{32.0} & \textbf{69.2} \\
\midrule
\multirow{3}{*}{[L]} & & & 51.9 & 23.3 & 69.8 \\
 & \checkmark & & - & 27.7 & - \\
 & \checkmark & \checkmark & \textbf{53.5} & \textbf{33.2} & \textbf{70.5} \\
\bottomrule
\end{tabular}
\smallskip\smallskip
\caption{Comparison on the baseline, scaling up during the inference procedure (denoted as \textbf{SU}), and fintuning with higher-resolution training images (denoted as \textbf{FT}). The baseline is the universal object detection training without either scaling up the input size when testing or the following dataset-specific high-resolution finetuning. In the table, [S] and [L] represent Large-UniDet~[S] and Large-UniDet~[L], respectively. The metric of COCO and MVD is mAP and the metric of OID is AP50.}\label{tab:up}
\end{table}

\begin{figure*}[!t]
  \centering
  \includegraphics[width=6.3in]{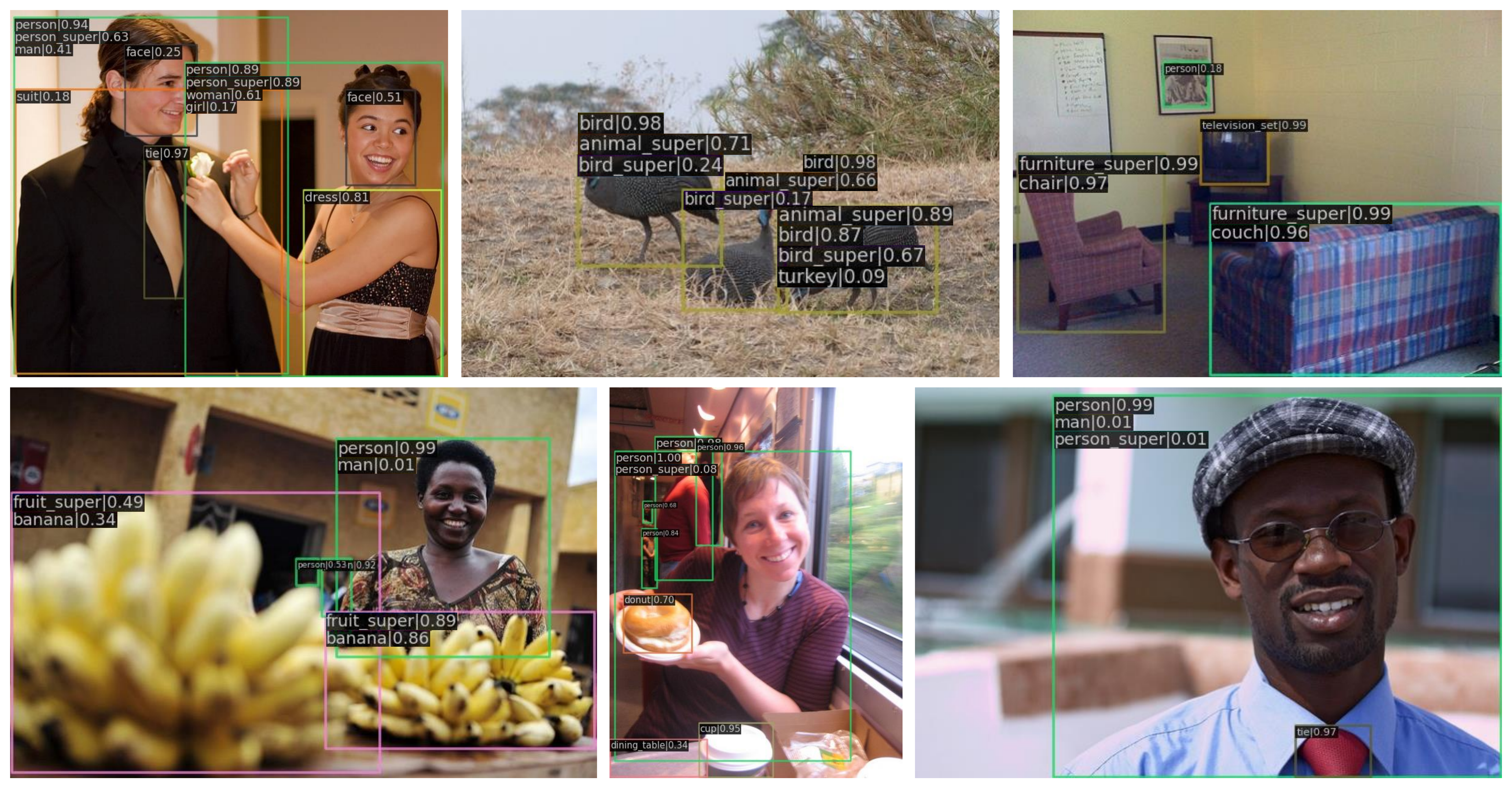}
  \caption{Qualitative results on COCO \emph{val} set. On the top left corner of each visualized bounding box, the predicted categories of which the corresponding confidence scores are greater than 0.01 are listed in the descending order of the confidence scores. The entry \emph{classname\_super} indicates this class is a superclass named \emph{classname} in the unified label space.}\label{fig:vis_coco}
\end{figure*}

\noindent
\textbullet~\emph{Dataset-specific finetuning with scaling up}

Employing an accelerated training approach involving high-resolution images subsequent to low-resolution pre-training yields commendable performance gains while curbing computational expenses~\citep{singh2018analysis}. Following the universal object detection training phase, we embark on dataset-specific fine-tuning at elevated resolutions. This decision is guided by a comprehensive consideration of performance metrics, training cost, and substantial distinctions inherent to the three datasets.

We use the cosine learning rate annealing with warm restarts~\citep{loshchilov2016sgdr} during the finetuning procedure.
Without any alterations to the model design, the finetuning and inference configurations are specified that,
\begin{itemize}
\item for COCO, the model undergoes 24 epochs of training with 6 cyclical restarts, while training images are scaled within a range of [640,~1200], and evaluation is performed using the 800-pixel resized short edge of testing images;
\item for MVD, the model undergoes 24 epochs of training with 6 cyclical restarts, while training images are scaled within a range of [1024,~2048], and evaluation is performed using the 2048-pixel resized short edge of testing images;
\item for OID, the model undergoes 6 epochs of training with 2 cyclical restarts, while training images are scaled within a range of [640,~1200] and evaluation is performed using the 800-pixel resized short edge of testing images.
\end{itemize}
Building upon the observed improvements resulting from inference scaling, we conduct finetuning with different hyper-parameters related to training image sizes. This optimization process aims to align the model more closely with the specific characteristics of the datasets. Notably, COCO and OID have relatively smaller raw image sizes compared to MVD. During universal object detection training, we originally considered training image sizes ranging from 480 to 960 pixels. However, in light of dataset-specific considerations, we empirically expand the range to [1024,~2048] for MVD training images, and [640,~1200] for COCO and OID. For a performance reason, we recommend dataset-specific finetuning to tailor the model to the unique characteristics of each dataset.

\begin{figure*}[!t]
  \centering
  \includegraphics[width=6.3in]{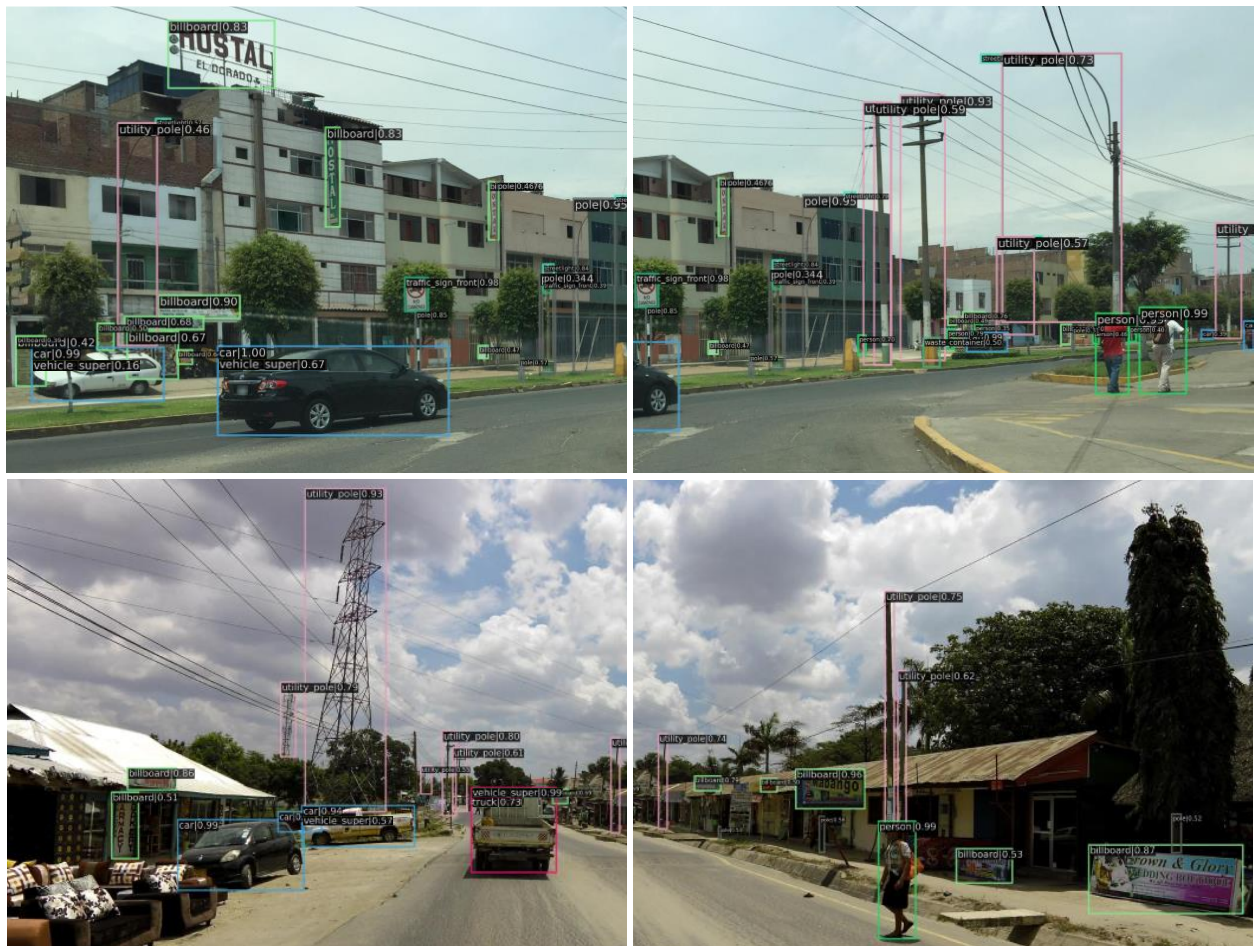}
  \caption{Qualitative results on MVD \emph{val} set. On the top left corner of each visualized bounding box, the predicted categories of which the corresponding confidence scores are greater than 0.1 are listed in the descending order of the confidence scores. The entry \emph{classname\_super} indicates this class is a superclass named \emph{classname} in the unified label space.}\label{fig:vis_mvd}
\end{figure*}

Table~\ref{tab:up} provides a summary of the results obtained through scaling and finetuning. With the lighter backbone, dataset-specific high-resolution finetuning yields significant improvements, including a 3.2 mAP increase on COCO, a substantial 10.9 mAP gain on MVD, and a 0.7 mAP improvement on OID. Even with the larger backbone, finetuning continues to deliver considerable enhancements across all three datasets, resulting in a +1.6 mAP improvement on COCO, a noteworthy +9.9 mAP increase on MVD, and a 0.7 mAP gain on OID, respectively.

\subsection{Visualization}
\begin{figure*}[!t]
  \centering
  \includegraphics[width=6.3in]{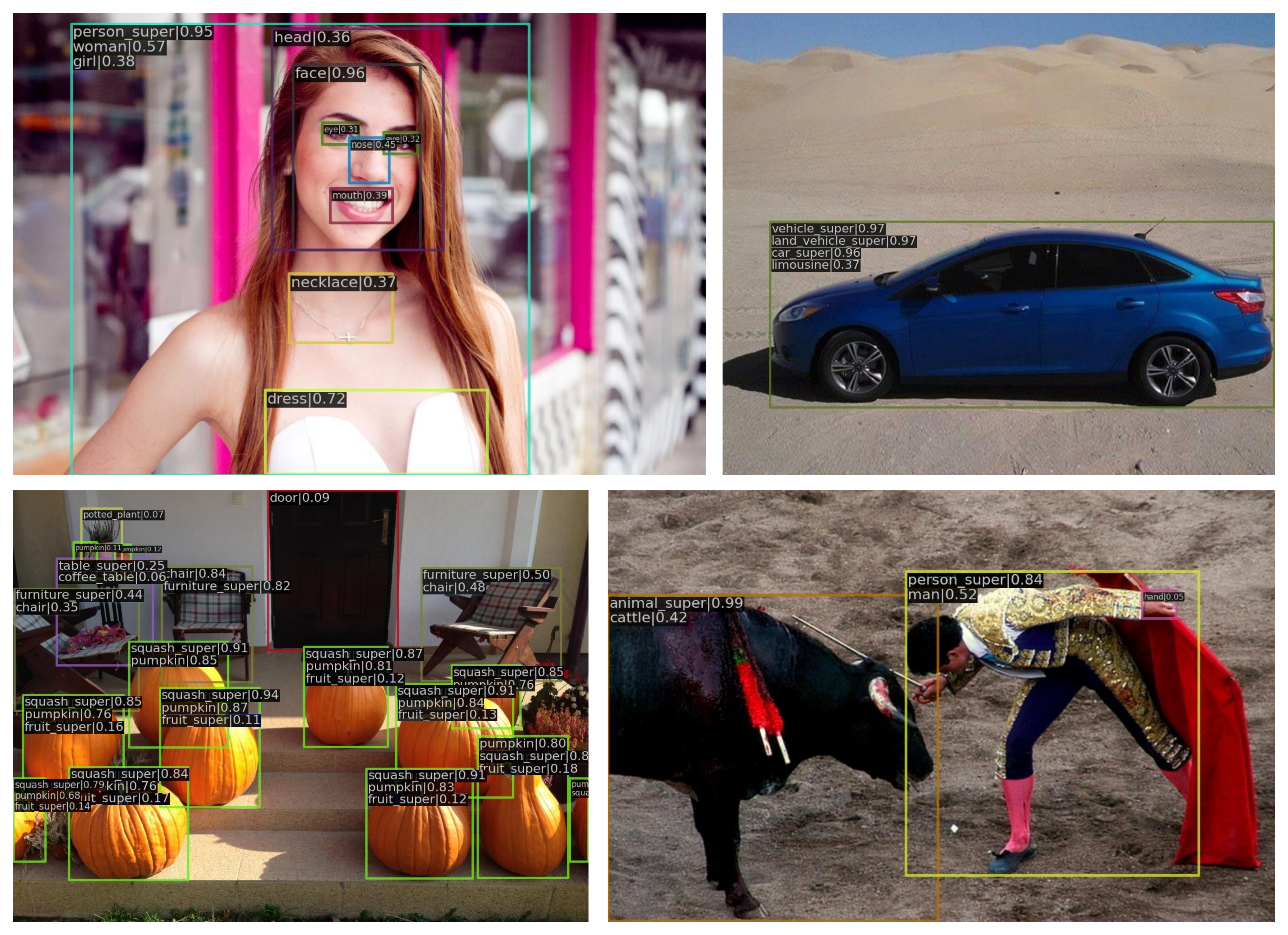}
  \caption{Qualitative results on OID \emph{val} set. On the top left corner of each visualized bounding box, the predicted categories of which the corresponding confidence scores are greater than 0.05 are listed in the descending order of the confidence scores. The entry \emph{classname\_super} indicates this class is a superclass named \emph{classname} in the unified label space.}\label{fig:vis_oid}
\end{figure*}

Figs.~\ref{fig:vis_coco},~\ref{fig:vis_mvd} and~\ref{fig:vis_oid} showcase the detection results of the Large-UniDet model which uses the SEER-RegNet256gf backbone, and has not undergone further high-resolution specific fine-tuning. These results display up to five recognized categories per bounding box with a confidence threshold, highlighting the presence of category label duplication and semantic hierarchy across datasets in universal object detection. We will delve into this phenomenon in greater detail and provide examples in the subsequent sections.

In the upper-left visualization result in Fig.~\ref{fig:vis_coco}, we can see that a significant number of the categories present in the original label space have been detected successfully. Additionally, some unannotated categories such as \emph{man}, \emph{woman}, \emph{girl}, \emph{person\_super}, \emph{face}, \emph{suit}, and \emph{dress} are also transferred from the other two datasets, despite not being part of the original label space in COCO.


\emph{Label duplication.} The categories \emph{person} and \emph{person\_super} exhibit semantic duplication, which results in high confidence scores for two individuals in this image. To illustrate, the man is classified as \emph{person} with a confidence score of 0.94 and \emph{person\_super} with a confidence score of 0.63. Similarly, the woman is classified as both \emph{person} and \emph{person\_super} with a confidence score of 0.89.

\emph{Semantic hierarchy.} As observed, since the categories \emph{woman}, \emph{man}, \emph{girl}, and \emph{boy} are all subclasses of \emph{person}, the two individuals in this image are classified under one or more of these four subclasses.

\section{Limitations}

\subsection{Cross-dataset Generalization}
In this study, we focus on the implementation of label space unification for three specific datasets, namely COCO, MVD, and OID. We acknowledge that generalizing this method to other diverse datasets, which may (a) lack hierarchical taxonomies, (b) contain different synonyms for similar concepts, or (c) utilize the same labels for distinct concepts, presents additional challenges and complexities that are not addressed within the scope of this study. 

To address these challenges and point toward potential solutions, we briefly offer several valuable approaches. (a) It's important to note that our method does not require a label hierarchy. Our loss strategy, HCLS, is not dependent on hierarchical taxonomies and can be applied without loss suppression for superclass categories. (b) To address the issue of different synonyms for similar concepts, one viable approach is to utilize a language model~\citep{raffel2020exploring} to align diverse synonyms and subsequently merge two categories representing similar concepts, as demonstrated in the literature~\citep{cai2022bigdetection}. (c) For situations involving the same labels being applied to distinct concepts, a plausible solution involves training a partitioned object detector with separate detection branches for different datasets. Subsequently, label spaces can be unified based on visual concepts, which aligns with strategies outlined in existing literature~\citep{zhou2022simple}.

Indeed, merging multi-domain taxonomies can be facilitated through the use of language models, pretrained object detectors, or a synergistic combination of both, presenting a promising pathway to address the challenge of label inconsistencies across diverse datasets. Beyond the above mentioned, in the following section, we delve into another issue stemming from label inconsistencies.

\subsection{Domain Over-fitting}
In Fig.~\ref{fig:vis_coco}, we observe that while the majority of annotated categories in the original COCO label space are successfully detected, some unannotated categories are also transferred from the other two datasets. However, accurately detecting and classifying these unannotated categories can be challenging due to annotation inconsistencies across the three datasets. For instance, in Fig.~\ref{fig:vis_coco}, categories like \emph{eye}, \emph{nose}, \emph{mouth}, and \emph{hand} from OID are not consistently detected in COCO images. Additionally, categories such as \emph{person\_super}, \emph{man}, \emph{woman}, and \emph{turkey}, which semantically match objects in COCO images, are either detected with low confidence scores or not detected at all. Similar results are observed in Fig.~\ref{fig:vis_mvd} and \ref{fig:vis_oid} for MVD and OID, where categories from other datasets may be omitted or exhibit varying levels of detection confidence.

Based on our observations, it appears that our universal object detector may be partially over-fitting to the distinctive characteristics of each data domain. This phenomenon suggests that the model has learned a way to distinguish testing images from different datasets, which could be seen as a form of ``\emph{cheating}'' in universal object detection. However, this behavior could pose challenges and potential harm in real-world practical applications, as it may lead to inaccurate or inconsistent results when deploying the model across diverse datasets.

\section{Discussion}


This raises an intriguing question: Can unifying the annotations in the unified label space lead to improved performance for the universal object detector on individual datasets? Existing literature~\citep{zhao2020object} has demonstrated the effectiveness of adding pseudo-labels for unannotated objects in the context of fully-annotated mixed datasets. However, the influence of annotation inconsistencies on individual datasets remains a relatively unexplored area.

In Section~\ref{losses}, we experimented with a similar label handling strategy called \emph{Unified hierarchy} to consolidate annotations for semantically duplicated categories. As we can see, our results presented in Tables~\ref{tab:label1} and \ref{tab:label2}, did not prove any improvement in COCO or MVD. Nevertheless, this topic warrants further exploration, and we leave it as an open question for future research.

\section{Conclusion}
In pursuit of solving the large-scale multi-domain universal object detection problem, we have introduced a series of resource-efficient techniques that leverage large vision models to yield robust visual representations across multiple diverse datasets. Our universal object detector combines three critical detector components with high capacity while maintaining the parameter stability of the large vision models. To address challenges such as cross-dataset label duplication and semantic hierarchy, we have implemented hierarchical taxonomy completion and a loss adaptation strategy known as HCLS within a unified label space spanning multiple datasets. Our research practices and findings offer a promising solution for real-world computer vision applications and underscore the potential of universal object detection in addressing complex, multi-domain challenges.

\backmatter
\section*{Data Availability Statement}
The authors confirm the data supporting the findings of this work are available within the article or its supplementary materials. 


\bibliography{sn-bibliography}



\clearpage
\onecolumn

\appendix
\section*{Appendix}
\section{RVC Submission}\label{simply}


We provide an in-depth presentation of our RVC final submission, which deviates slightly from the configuration outlined in the main text. Our final RVC submission employed a customized variant of the proposed detector built upon SEER-RegNet256gf. To align with the RVC deadline, specific simplifications were incorporated into the model:
\begin{itemize}
\item Instead of utilizing the default setting of {$C_2$, $C_3$, $C_4$, $C_5$} in NAS-FPN, we opted to employ the side-outputs {$C_3$, $C_4$, $C_5$} and applied a $2\times$ downsampling on C5 twice, resulting in a 5-level feature pyramid. While this simplification led to a reduction in accuracy for detecting small objects, it significantly accelerated the training process.
\item The basic anchor scale in Cascade RPN was reduced to 5.04 ($4\times2^{1/3}$). This adjustment was made to align with the changes in NAS-FPN and to minimize instances of missed detections for small objects.
\item The model underwent training for 720k iterations, with a learning rate reduction of 0.1 applied at 600k iterations.
\end{itemize}
During the dataset-agnostic inference procedure, the Soft-NMS~\citep{bodla2017soft} was performed with an IoU threshold of 0.6 and a score threshold of 0.001. Then,
\begin{itemize}
\item for COCO, we limited the maximum number of predictions per image to 100, and the short edge of the input image was resized to 800 pixels;
\item for MVD, we limited the maximum number of predictions per image to 300, and the short edge of the input image was resized to 2048 pixels;
\item for OID, we limited the maximum number of predictions per image to 300, and the short edge of the input image was resized to 800 pixels.
\end{itemize}
We did \textbf{not} employ any advanced inference techniques, such as multi-scale test augmentation. The performance of our submission (IFFF\_RVC) on the three datasets is summarized in Table~\ref{tab:result}. For comparison with the results presented in this paper, we evaluated the model for our RVC submission on the validation sets with a maximum of 300 predictions per image, using the standard Non-Maximum Suppression (NMS) method. All other testing configurations remained consistent with those used on the test sets.

\end{document}